\documentclass{article}
\usepackage{spconf,amsmath,graphicx}
\usepackage{mathtools}
\usepackage{bbold}
\usepackage{algorithm} 
\usepackage{algpseudocode, setspace}
\usepackage{siunitx}
\usepackage[center]{subfigure}
\usepackage{graphicx,float}
\usepackage[table]{xcolor}
\usepackage[numbers]{natbib}
\definecolor{gray}{rgb}{0.85,0.85,0.85}
\usepackage{tabularray}
\usepackage{booktabs,arydshln}
\usepackage{xcolor}
\usepackage[export]{adjustbox}
\usepackage{hyperref}
\usepackage{cleveref}
\usepackage{comment}
\hypersetup{
	colorlinks   = true,    % Colours links instead of ugly boxes
	urlcolor     = blue,    % Colour for external hyperlinks
	linkcolor    = red,     % Colour of internal links
	citecolor    = green    % Colour of citations
}

\graphicspath{{IMAGES/}}

\definecolor{dg}{rgb}{0.0, 0.5, 0.0}

\title{A study on altering the latent space of pretrained Text to Speech Models for improved expressiveness}
\name{Mathias Vogel \thanks{Work done as part of an internship at \href{https://mtc.ethz.ch/}{ETHZ MTC} }}
\address{Media Technology Center, ETH Z\"urich, Switzerland}

\begin{document}
%\ninept
\maketitle
\begin{abstract}
This report explores the challenge of enhancing expressiveness control in Text-to-Speech (TTS) models by augmenting a frozen pretrained model with a Diffusion Model that is conditioned on joint semantic audio/text embeddings. The paper identifies the challenges encountered when working with a VAE-based TTS model and evaluates different image-to-image methods for altering latent speech features. Our results offer valuable insights into the complexities of adding expressiveness control to TTS systems and open avenues for future research in this direction.
\end{abstract}

\begin{keywords}
Machine Learning, Signal Processing, Text-to-Speech Synthesis
\end{keywords}

%\noindent\let\thefootnote\relax\footnotetext{\url{https://github.com/majedelhelou/FC-Diffusion}}

\newcommand{\ZT}{$Z_{\text{text}}$}
\newcommand{\ZA}{$Z_{\text{audio}}$}
\newcommand{\hZA}{$\hat{Z}_{\text{audio}}$}

\section{Introduction}\label{sec:intro}
Significant progress has been made in the development of Text-to-Speech (TTS) systems, with models such as VITS \cite{Kim2021ConditionalText-to-Speech} achieving a mean opinion score comparable to that of genuine speech recordings. However, the difficulty of having precise control over the prosodic features of generated speech samples remains an unsolved issue. Many TTS models lack mechanisms to control prosodic and emotional nuances, which are essential for a wide range of applications.

In this paper, we present an exploratory study in which we enhance the VITS model with expressiveness control by adding a Denoising Diffusion Model (DDM) \cite{Ho2020DenoisingModels} conditioned on joint audio/text embeddings such as CLAP embeddings \cite{Wu2023Large-ScaleAugmentation} to alter the latent VITS encodings. We chose DDM because these models are known to be easy to condition and are currently state-of-the-art in many computer vision tasks \cite{Ho2022VideoModels, Hoogeboom2023SimpleImages, Rombach2021High-ResolutionModels}. The final goal is to be able to change the generated speech by providing a target style by either providing a recording by text promts describing the style. Contrary to our expectations, the method did not produce the desired improvements in expressiveness control. However, we believe that the findings of our study offer valuable insight into the complexities and challenges associated with adding expressiveness control to TTS systems. Our findings could help future research in designing systems that allow greater control over prosodic and emotional features.

Our contributions are as follows:
\begin{enumerate}
    \item We identify challenges of working with pretrained VAE based TTS models.
    \item We apply and compare different image-to-image methods to change latent speech features, highlighting their strengths and weaknesses.
    \item We open up discussion on further research directions of controlled emotional TTS.
\end{enumerate}

\section{Method Preliminary}
\begin{figure}[ht!]
    \centering
    \includegraphics[width=\linewidth]{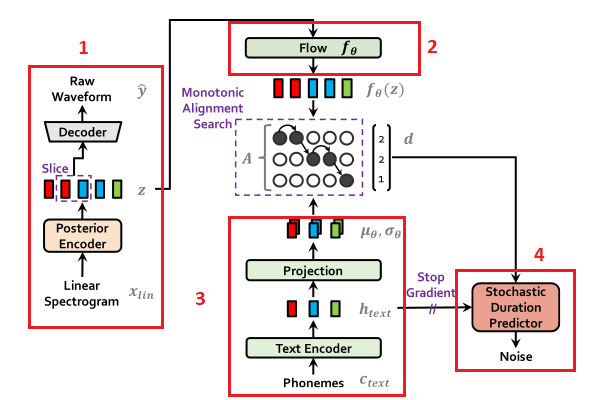}
    \caption{The training procedure of VITS.}
    \label{fig:vits_training}
\end{figure}

We chose the VITS model as our backbone TTS model due to its fast inference speed and quality. This model is trained using adversarial learning, as illustrated in \cref{fig:vits_training}. The architecture consists of several parts, namely:

\begin{enumerate}
    \item A conditional variational auto-encoder (CVAE) with a WaveGlow \cite{Prenger2019Waveglow:Synthesis} based encoder and a HiFi-GAN \cite{Kong2020HiFi-GAN:Synthesis} based decoder acting as a neural vocoder. The CVAE embeds a linear spectrogram $x_{lin}$ in a lower dimensional space $\text{Z}$ to lower the computational costs of the model.
    \item A normalizing flow that provides an invertible mapping between the complex distribution $\text{Z}$ and the simpler distribution $f_\theta(\text{Z})$.
    \item A text encoder that processes input phonemes $c_{text}$ to statistics $\mu_{\theta}$ and $\sigma_{\theta}$ that represent the probabilities of learned representations $h_{text}$ given the input phonemes and the spectrogram $x_{lin}$ of the corresponding audio.
    \item A stochastic duration predictor that learns the length of the phoneme representations.
\end{enumerate}

\begin{figure}[ht!]
    \centering
    \includegraphics[width=.5\linewidth]{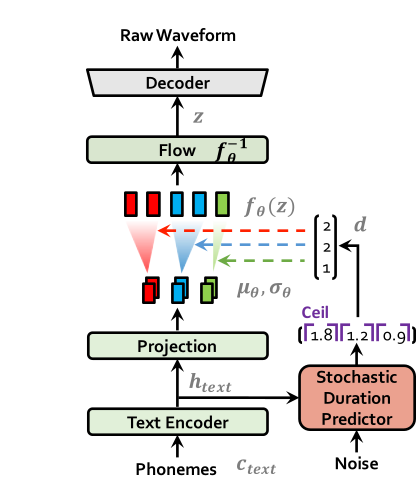}
    \caption{VITS inference.}
    \label{fig:vits_inference}
\end{figure}

\Cref{fig:vits_inference} illustrates the use of VITS during inference, where the model does not require the posterior encoder component of the CVAE and directly transforms a text input into speech.

\section{Method}
\begin{figure}[ht!]
    \centering
    \includegraphics[width=\linewidth]{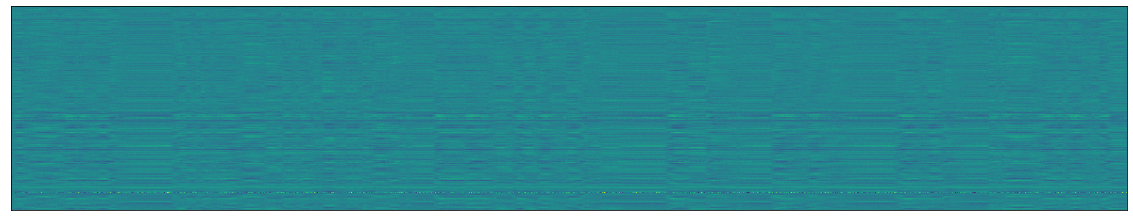}
    \caption{Visualization of a latent sample $Z$ which reminds of a spectrogram.}
    \label{fig:z_spec}
\end{figure}

In our proposed method, we alter the CVAE bottleneck embeddings $Z$ of VITS. Because samples of $Z$ resemble a spectrogram \cref{fig:z_spec}, we apply approaches from Image-to-Image translation.

The goal is to transform a neutral $Z$ into a stylized $Z'$ while preserving the content. To be able to control the style, we condition the diffusion process on semantic text-audio embeddings $c$ obtained using CLAP. CLAP is trained using a contrastive loss, which should result in embeddings that are close together for audio recordings and text prompts that match semantically, such as a recording of someone shouting and the phrase "person shouting". Since there are no semantic descriptions available for the LJS \cite{Ito2017TheDataset} and VCTK \cite{Veaux2017CSTRToolkit} datasets on which VITS was pre-trained, we use embeddings of the audio only during training. However, a data set with style annotation would be very beneficial.

\begin{figure}[ht!]
    \centering
    \includegraphics[width=.5\linewidth]{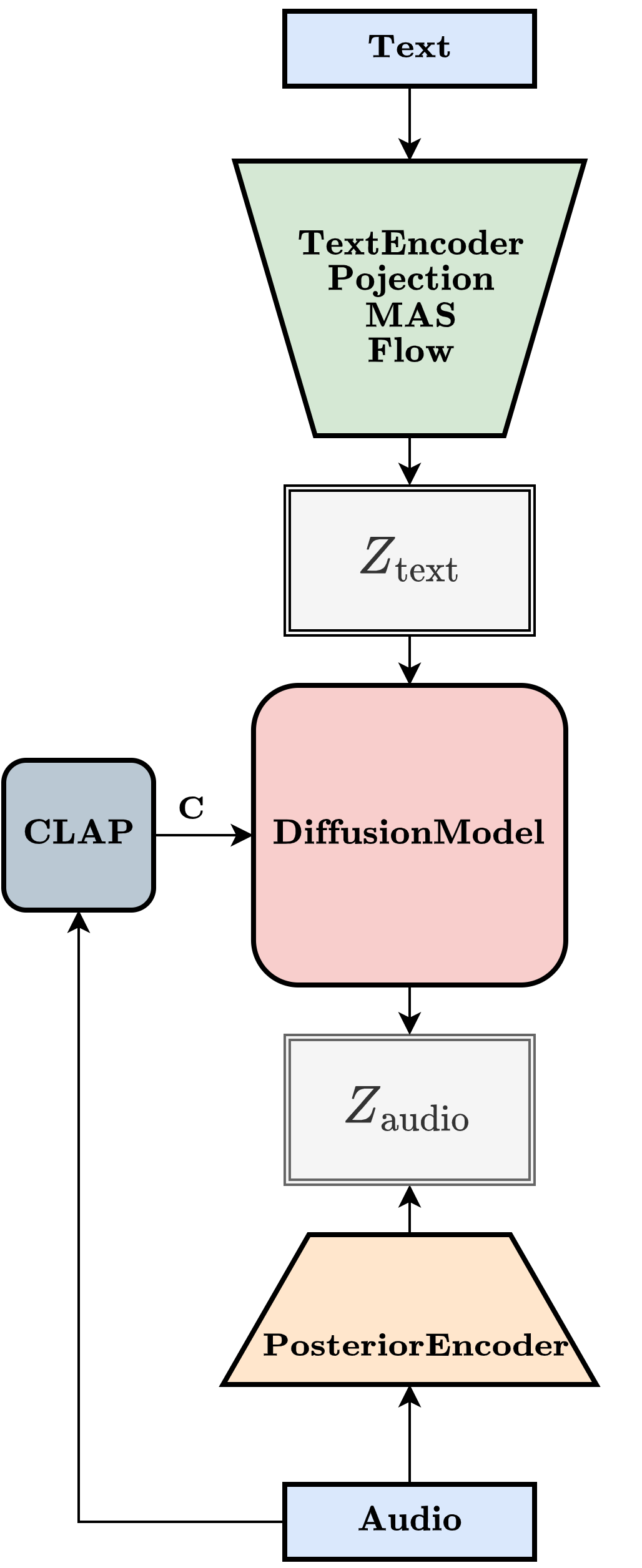}
    \caption{Training of the proposed method.}
    \label{fig:method_sketch}
\end{figure}

We obtain training data for $Z$ and $Z'$ using different modules of VITS. To generate $Z$, we simply apply the frozen VITS text encoder to the transcript of a speech recording. We denote the resulting embeddings by \ZT. Encoding the corresponding speech recording with the VITS \emph{PosteriorEncoder} we obtain \ZA~embeddings. Because \ZA~is produced from a speech recording, it should contain more prosodic features than \ZT~ since \ZT~ just matches the average style of the speaker. Both \ZA~and \ZT~are of shape [C, H, W], where zero padding is applied to the width dimension when needed. \Cref{fig:method_sketch} shows a complete sketch of our training method.

\section{Experimental Evaluation}
\label{sec:experiments}

The resulting audio of all the experiments reported can be heard on \href{https://github.com/MrBirrd/natural_voices_demo}{GitHub}. We divide the results by their Image-to-Image modeling approach. Independent of the approach, we train our models using the Adam optimizer, a constant learning rate of 1e-4, and batch size of 64. We train all our models for around 50k steps using the mean-squared error loss. We use the v-objective introduced in \cite{Salimans2022PROGRESSIVEMODELS} and DDIM sampling \cite{Song2021DENOISINGMODELS} using 10 sampling steps, which was experimentally found to be a good compromise between speed and quality.

\subsection{Palette}
\label{sec:palette}
For our baseline we adapt the Palette \cite{Saharia2022Palette:Models} Image-to-Image diffusion model for our task by allowing the U-Net \cite{Ronneberger2015U-net:Segmentation} backbone, inspired by recent works \cite{Dhariwal2021DiffusionSynthesis, Nichol2021ImprovedModels}, to process non-squared data effectively. We add the information from \ZT~by channel-wise concatenation with diffusion noise $\epsilon$ for each diffusion timestep such that the input of the model has a shape of [B, 2C, H, W]. Conditioning on the CLAP embedding $c$ is implemented through cross-attention following \cite{Rombach2021High-ResolutionModels}.
\\

The resulting speech samples using this method do not contain any intelligible content, although there is a certain melody and style to them. The potential cause of this could be that the model is only able to concentrate on style alterations when the content of the source \ZT~and the target \ZA~are the same. However, due to the stochastic nature of VITS, specifically its \emph{Stochastic Duration Predictor}, the size of the source and target latent spectrogram is not the same and sometimes differs by as much as $50\%$. This could potentially lead the training objective to consist mainly of misalignment errors rather than differences in style. We hypothesize that conditioning by channel-wise concatenation does not lead to good results, as this approach originates from image super-resolution \cite{Saharia2023ImageRefinement, Ho2022CascadedGeneration}, where the underlying content is aligned by definition. Instead, the content could be provided to the model using a pre-trained model such as Wav2Vec \cite{Baevski2020Wav2vecRepresentations} and concatenate the extracted embeddings with the CLAP embeddings.

However, even if we were capable of providing the content and target style to the diffusion model in an effective manner, there is no assurance that the final audio generated by the altered VITS still matches our target style. This is because the VITS decoder is trained as a person-specific vocoder, which could remove potential style changes and produce the same average person-specific speech. This hypothesis is tested in the following experiment, where we also explore a more advanced Image-to-Image method.

\subsection{I2SB}
\begin{figure}[h]
    \centering
    \includegraphics[width=\linewidth]{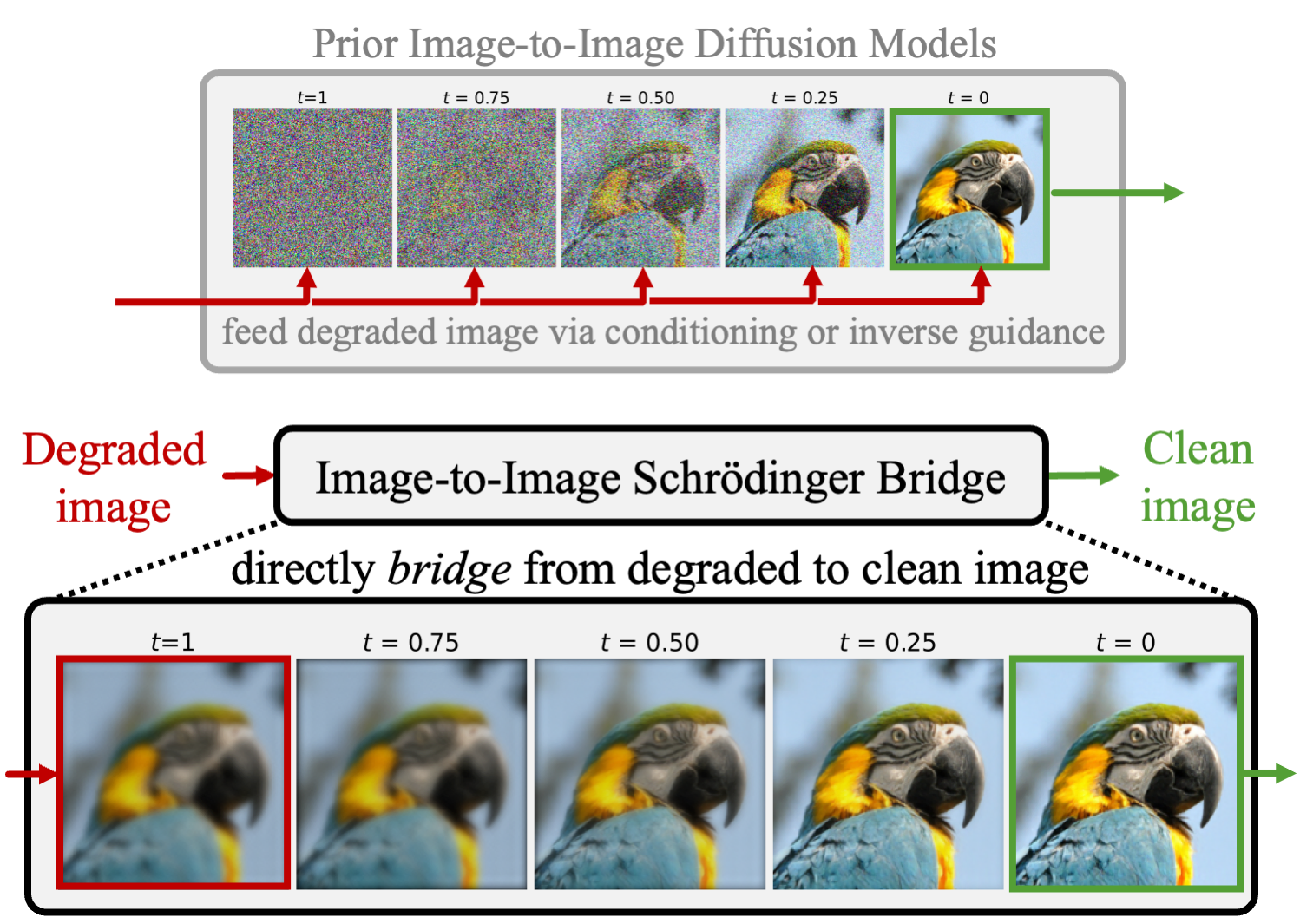}
    \caption{Comparison of a traditional image-to-image translation task shown on top with the I2SB method, where a direct bridge between the initial distribution and the target distribution is learned.}
    \label{fig:i2sb}
\end{figure}

I2SB stands for Image-to-Image Schrödingers Bridge \cite{Liu2023I2SB:Bridge} and is capable of producing state-of-the-art results in many image-to-image tasks such as deblurring, (freeform) inpainting or super-resolution. In contrast to the Palette diffusion process, which maps isotropic noise to a sample by conditioning on an image via concatenation, I2SB does not rely on concatenation and directly maps one distribution $p_0(x)$ into another distribution $p_1(x)$. This process, as well as the difference to approaches such as Palette, is visualized in \cref{fig:i2sb}. I2SB allows for more general image-to-image translation tasks by adapting the way the bridge is constructed and the way to condition the process.

We use the same U-Net backbone as described in \cref{sec:palette} for an I2SB pipeline using the \verb|--add-x1-noise| and \verb|--ot-ode| options. These settings are suggested by the authors of I2SB when dealing with more challenging tasks where the two distributions $p_0(x)$ and $p_1(x)$ are potentially very different from each other. We could validate the suggestions experimentally.
\\

The resulting speech samples using the I2SB approach are qualitatively similar to the Palette approach. The samples contain little noise, but the content is inaudible, which can be heard on our project GitHub. We also evaluate the impact of Classifier-Free Guidance (CFG) \cite{Ho2021Classifier-FreeGuidance} which shows that a higher guidance scale changes both the content and the style, however, only the style is expected to change. This experiment shows that our models use the CLAP embedding in a different way than intended. This could be due to the aforementioned lack of content alignment between \ZT~and \ZA~which leads to the model trying to extract missing content from the CLAP embeddings instead of extracting only prosodic speech features.

In addition to changing the CFG scale, we also conducted an experiment to understand the impact of the VITS vocoder. For this experiment, we used ground truth samples from two different speakers A and B of the VCTK test data set. The samples have a different content. We denote ground-truth audio recordings as $\text{Audio}_{A}$ and $\text{Audio}_{B}$ with their corresponding transcripts $\text{T}_{A}$ and $\text{T}_{B}$ and CLAP audio embeddings $\text{CLAP}_{A}$ and $\text{CLAP}_{B}$. If we want to recreate speaker A we would condition the diffusion model on $\text{CLAP}_{A}$ and set the speaker id to A in the VITS vocoder. On the other hand, if the goal is to obtain a speech sample of speaker A in the style of speaker B, we would condition the diffusion model on $\text{CLAP}_{B}$ while conditioning the VITS vocoder on the id of A. By conditioning the diffusion model on $\text{CLAP}_{B}$ and the VITS vocoder on the id of B, we should obtain a speech sample that sounds exactly like speaker B.

The samples in our \href{https://github.com/MrBirrd/natural_voices_demo}{GitHub} demonstrate that this approach does not work as expected. Conditioning the diffusion model on a CLAP embedding of a different speaker changes the content more than the style, while changing the conditioning id of the VITS vocoder changes the style significantly. These findings suggest that an approach that aims to change the latent space Z of a VAE to achieve style change can not work as intended due to the fact that the decoder is conditioned on a speaker id and that the latent space Z contains mostly information about the speech content instead of style information.

\section{Conclusion} \label{sec:ccl}
In this report, we conducted an in-depth examination of adding expressiveness control to Text-to-Speech (TTS) systems, specifically focusing on the VITS model. Our approach aimed to modify the latent embeddings (Z) of a frozen VITS model using a Diffusion Model conditioned on joint audio/text embeddings like CLAP. The goal was to offer a mechanism for controlling the expressiveness of speech by providing a speech sample of reference style or a text prompt describing the desired style.

Contrary to our expectations, our method did not produce significant improvements in expressiveness control. Despite this, the research provides several insights into the complexities and challenges related to augmenting TTS systems with expressiveness control features.

Our contributions can be summarized as follows:
\begin{enumerate}
    \item We identified challenges tied to working with pretrained VAE-based TTS models like VITS, including the limitations of their latent spaces in encoding stylistic features.
    \item We applied and evaluated different image-to-image translation methods, including Palette and I2SB, to alter latent speech features, thus revealing their strengths and weaknesses.
    \item We opened the door to future research directions in the field of controlled emotional TTS by outlining the complexities involved.
\end{enumerate}

The most significant limitation encountered was the mismatch in content alignment between \ZT~and \ZA, which prevented our diffusion models from effectively altering only the style characteristics. Our experiments also demonstrated that the speaker-specific conditioning of the VITS vocoder significantly impacts the stylistic outcome, raising questions about the feasibility of changing the style via the latent space Z.

Given these findings, future research could explore alternative methods to modify the latent spaces of TTS models. This could involve the use of separate and disentangled latent spaces for style and content. Another approach could focus on directly training a style-conditional diffusion model to map from a style-neutral latent space Z to speech audio.

\bibliographystyle{IEEEbib}
\bibliography{main}

\end{document}